\documentclass[10pt,twocolumn,letterpaper]{article}

\usepackage{cvpr}              % To produce the CAMERA-READY version
\usepackage[dvipsnames]{xcolor}

\usepackage{times}
\usepackage{epsfig}
\usepackage{graphicx}
\usepackage{amsmath}
\usepackage{amssymb}
\usepackage{color}
\usepackage{multirow}
\usepackage{setspace}
\usepackage{chngcntr}
\usepackage{float}
\usepackage[accsupp]{axessibility}
\usepackage{multirow}
\usepackage{hhline}
\usepackage{colortbl}
\definecolor{myorange}{RGB}{255,165,0}
\definecolor{mygray}{gray}{0.7}
\usepackage{caption}
\definecolor{cvprblue}{rgb}{0.21,0.49,0.74}
\usepackage[pagebackref,breaklinks,colorlinks,citecolor=cvprblue]{hyperref}

\def\paperID{9899} % *** Enter the Paper ID here
\def\confName{CVPR}
\def\confYear{2024}

\title{LoSh: Long-Short Text Joint Prediction Network for Referring Video Object Segmentation}

\author{Linfeng Yuan,~~Miaojing Shi\thanks{Corresponding author.},~~Zijie Yue,~~Qijun Chen\\
College of Electronic and Information Engineering, Tongji University\\
{\tt\small linfengyuan1997@gmail.com, \{mshi,zijie,qjchen\}@tongji.edu.cn}
}

\begin{document}
\maketitle

%%%%%%%%% ABSTRACT
\begin{abstract}
Referring video object segmentation (RVOS) aims to segment the target instance referred by a given text expression in a video clip. The text expression normally contains sophisticated description of the instance's appearance, action, and relation with others. It is therefore rather difficult for a RVOS model to capture all these attributes correspondingly in the video; 
%\textcolor{red}{in fact, the model's ineffectiveness in failure cases often results from its excessive bias towards the action- and relation-related visual attributes of the instance.}
in fact, the model often favours more on the action- and relation-related visual attributes of the instance. 
This can end up with partial or even incorrect mask prediction of the target instance. We tackle this problem by taking a subject-centric short text expression from the original long text expression. The short one retains only the appearance-related information of the target instance so that we can use it to focus the model's attention on the instance's appearance. We let the model make joint predictions using both long and short text expressions; and insert a long-short cross-attention module to interact the joint features and a long-short predictions intersection loss to regulate the joint predictions. Besides the improvement on the linguistic part, we also introduce a forward-backward visual consistency loss, which utilizes optical flows to warp visual features between the annotated frames and their temporal neighbors for consistency.    
%propose a general training approach. 
%It is built upon the current query-based RVOS frameworks, termed \textbf{L}ong-\textbf{S}hort Text Joint \textbf{Pred}iction Network (LoSh). 
%Our framework extracts the subject-centric part of a long text expression as a short one and inputs them together into the linguistic encoder. Previous RVOS models take complete text expressions as input and generate exactly or partially wrong masks in many cases. This stems from the fact that the complex structure of long text expressions makes models pay excessive attention to the action or relation, rather than the appearance of the target instance. To alleviate this problem, we introduce the short text expressions whose semantics are closer to the appearance of the referred instance than its action or relation. 
%We propose a novel long-short prediction intersection loss between the predictions for long and short text expressions. It utilizes predictions for appearance-focused short text expressions to align or penalize the predicted pixels for action-focused long ones. 
%This mechanism greatly reduces instance-level misclassification and significantly boosts segmentation performance. 
We build our method on top of two state of the art pipelines. 
Extensive experiments on A2D-Sentences, Refer-YouTube-VOS, JHMDB-Sentences and Refer-DAVIS17 show impressive improvements of our method.
Code is available \href{https://github.com/LinfengYuan1997/LoSh}{here}.
%show that our method LoSh built upon the state-of-the-art methods gains significant improvements without any modification of network structures. Without extra training data and parameters, LoSh-M shows 
\end{abstract}    
\section{Introduction}

%\textcolor{red}{Reminder for Linfeng and Miaojing:
%1) Experiments on the Refer-YouTube-VOS dataset should be done later.
%2) Update the experiment results on ReferFormer after getting stable results.}

\begin{figure}[t]
\begin{center}
   \includegraphics[width=0.9\linewidth]{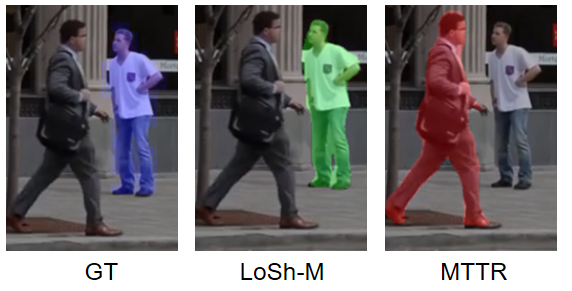}
\end{center}
\setlength{\abovecaptionskip}{-0.3cm}
\setlength{\belowcaptionskip}{-0.3cm}
\caption{Qualitative comparison between our LoSh-M and MTTR~\cite{MTTR}. 
%From upper left to lower right, they are the center frame of the input video, mask annotations, the output of LoSh, and the output of MTTR. 
The text expression for the target instance is `a man in white t-shirt is walking'. LoSh-M generates an accurate  prediction while MTTR predicts a wrong mask compared to ground truth.} 
%Our proposed LoSh generates prediction covering the referred instance while MTTR generates a completely wrong mask.}
\vspace{-6pt}
\label{fig:long}
\label{fig:onecol}
\end{figure}

Referring video object segmentation (RVOS) \cite{firstRVOS} aims to segment the target instance in a video given a text expression, which can potentially benefit many video applications such as video editing and surveillance. For the research community, RVOS is a challenging and interesting multi-modal task: videos provide dense visual information while text expressions provide symbolic and structured linguistic information. This makes the alignment of the two modalities very challenging, especially for the segmentation task. A similar task to RVOS is the referring image segmentation (RIS) \cite{RIS2, RIS1,zhou2024text}, which aims to segment the target instance in an image given a text expression. Compared to RIS, RVOS is significantly harder due to the difficulties to tackle the motion blur and occlusion in video frames. It is essential to build the data association across multiple frames and track the target instance in the video. Besides the visual difference between RIS and RVOS, the text expressions in them are also different: those in RIS mainly describe the appearance of the target instances in static images while those in RVOS describe both the appearance and action of the target instances over several frames in videos. Therefore, a robust RVOS model is expected to capture complex textual information from the input text expression and correspondingly align it with visual features in video frames.

Recently, the development of transformer in video feature extraction~\cite{videoswint}  and instance segmentation~\cite{VisTR} benefits RVOS a lot. Especially, MTTR~\cite{MTTR} and ReferFormer~\cite{ReferFormer} are pioneering works to adapt query-based transformer models~\cite{DETR, DeformableDETR, VisTR} to RVOS and have achieved significant improvements compared with previous methods. 
The performance gains in most query-based RVOS frameworks mainly come from the strong capabilities of transformer on feature extraction and multi-modal feature fusion  while the exploitation of the linguistic part has not been emphasized.

%due to the strong abilities to capture spatial-temporal visual features and align these visual features with linguistic ones.\linfeng{It highlights that a large part of the improvement of the Transformer-based RVOS method comes from the mining of visual features.} Although recent RVOS frameworks have gained promising performance, most of them do not emphasize the potential of exploration in the language aspect.

%\linfeng{Problem of previous work}
The text expression for the target instance normally contains a sophisticated description of the instance (subject)'s appearance, action, and relation with others. Capturing all these attributes in a video presents a significant challenge for an RVOS model. 
{We have analyzed the failure predictions by a state of the art model, MTTR~\cite{MTTR}, on the A2D-Sentences~\cite{firstRVOS} dataset, whose  IoUs with GTs are less than 0.5: we randomly sample 400 such cases and observe that over 70\% of them either mis-align with appearance-related  phrases or overly concentrate on the discriminative regions corresponding to actions or relations.}
% We observe that
This suggests that the model's mask prediction tends to favour more on the instance's action or relation with others rather than its appearance. This can lead to partial segmentation focusing on the action-related part of the target instance, or incorrect segmentation if the prediction mistakenly focuses on another instance that behaves similarly to the target instance.
Fig.~\ref{fig:long} illustrates an example of this: the video frame contains two men on the street. Given the text expression, `a man in white t-shirt is walking', it refers to the man in a white t-shirt by the right of the frame. However, we run  
% a state of the art query-based RVOS model, 
MTTR~\cite{MTTR} on this video and observe that its prediction instead covers the walking man in a gray suit by the left of the frame. Apparently, MTTR has favoured more on the word `walking' rather than `white t-shirt' in the text expression.

%\linfeng{motivation of long-short and calculation of the intersection loss}
To tackle this problem, our essential idea is to reduce the excessive impact of action/relation-related expression on the final mask prediction. We can first generate a subject-centric short text expression (\eg, `a man in white t-shirt') by removing the predicate- and object-related textual information from the original long text expression (\eg, `a man in white t-shirt is walking'). The short 
 text expression should contain only the subject and the adjective/phrase that describes the subject.  
%from the complete text expressions as short ones. 
It is a more general expression for the referred instance compared to the long one.  Given the input video, the mask prediction for the short text expression pays attention to the instance's appearance; while the mask prediction for the long text expression pays attention to both the instance's appearance and action, though the latter is often more favored. Ideally, the latter mask prediction should be included in the former. To take advantage of the relations between long and short expressions, in the feature level, we introduce a long-short cross-attention module to strengthen the feature corresponding to the long text expression via that from the short one; in the mask level,   
%We propose a long-short cross-attention mechanism by utilizing alternating self-attention and long-short cross-attention layers instead of consecutive self-attention layers in the original transformer encoder.  
we introduce a long-short predictions intersection loss to regulate the model predictions for the long and short text expressions.

Apart from improving the RVOS model on the linguistic part, 
we also target to improve the model on the visual part by exploiting temporal consistency over visual features. 
Previous methods~\cite{AAMN,modelingmotion} in RVOS assume the availability of optical flows between video frames and utilize them to generate auxiliary visual features to enhance video features.   
%for  extract features of the optical flow by convolutional layers for better video understanding \miaojing{not okay}. 
Different from them, we introduce a forward-backward visual consistency loss  to directly compute optical flows between video frames and use them to warp the features of every annotated frame's temporal neighbors to the annotated frame for consistency optimization. 
%Our optic flows are generated \miaojing{xxx}  
%a vision-based innovation, which can advance the temporal consistency of pixels in the feature maps across multiple frames. As mentioned above, the data association across multiple frames is crucial for RVOS task.

% Inspired by~\cite{EFC} in the video segmentation task, which takes two frames as input and warp the feature \miaojing{also the feature?} from one to the other via optical flow,  \miaojing{any difference?}

%\linfeng{briefly introduce the consistency in EFC}
%And the optical flows across frames can help the RVOS model capture the temporal information. Given this motivation, we adapt the warping consistency loss in \cite{EFC} to RVOS and 

%to strengthen feature-level consistency across frames in the latent space. Specifically, we calculate the optical flows from the keyframe to adjacent frames by \cite{DenseOpticalFlowEstimation} which does not require training. By these optical flows, we can compute the warped features of the keyframe from adjacent frames. Finally, minimizing the $l_2$ distances between features of the keyframe and warped ones can improve the temporal feature extraction ability of the visual encoder.

Overall, our main contribution is we develop a long-short text joint prediction network (LoSh) to segment the referred target instance under properly designed guidance from both long and short text expressions. Two components are emphasized, the long-short cross-attention module and long-short predictions intersection loss.
%can aggregate useful information from multi-modal features corresponding to short text expressions to those features for long ones. In terms of network optimization, 2) we propose a long-short predictions intersection loss to constrain the model to generate more consistent mask predictions for long and short text expressions.
Besides, we inject a forward-backward visual consistency loss into LoSh to  exploit the feature-level temporal consistency between temporally adjacent visual features.

We evaluate our method on standard RVOS benchmarks, \ie, A2D-Sentences~\cite{firstRVOS}, Refer-YouTube-VOS~\cite{Urvos}, JHMDB-Sentences~\cite{firstRVOS} and Refer-DAVIS17~\cite{refdavis17} datasets. We show that LoSh significantly outperforms state of the art across all metrics.

\section{Related work}
\noindent
\textbf{Referring video object segmentation.} %The goal of RVOS task is to generate a  pixel-wise mask prediction for the target instance in a video using a text expression. 
%The text expression contains descriptions regarding the target instance's appearance, action, or relation to others. 
% Gavrilyuk \etal ~\cite{firstRVOS} introduce this task and propose an encoder-decoder structure to generate the segmentation mask by convolving visual features with dynamic filters obtained from linguistic features.
Previous RVOS methods are realized either in a single-stage or two-stage manner. Single-stage methods~\cite{Urvos, RefVOS, CSTM, CMSA} directly fuse the visual and linguistic features extracted from the input video and text expression; and generate the final mask prediction on top of the fused features with a pixel decoder. In contrast, two-stage methods\cite{AAMN, rethinking, clawcranenet}  first generate a number of instance candidates in the video based on the visual features; then select the one that has the highest matching score with the input text expression as the final prediction. In general, two-stage methods perform better than single-stage ones. However, they also suffer from heavy workloads compared to single-stage ones. 

\begin{figure*}
\begin{center}
% \fbox{\rule{0pt}{2in} \rule{.9\linewidth}{0pt}}
% \includegraphics[width=0.8\linewidth]
% {overview}
\includegraphics[width=0.9\linewidth]
{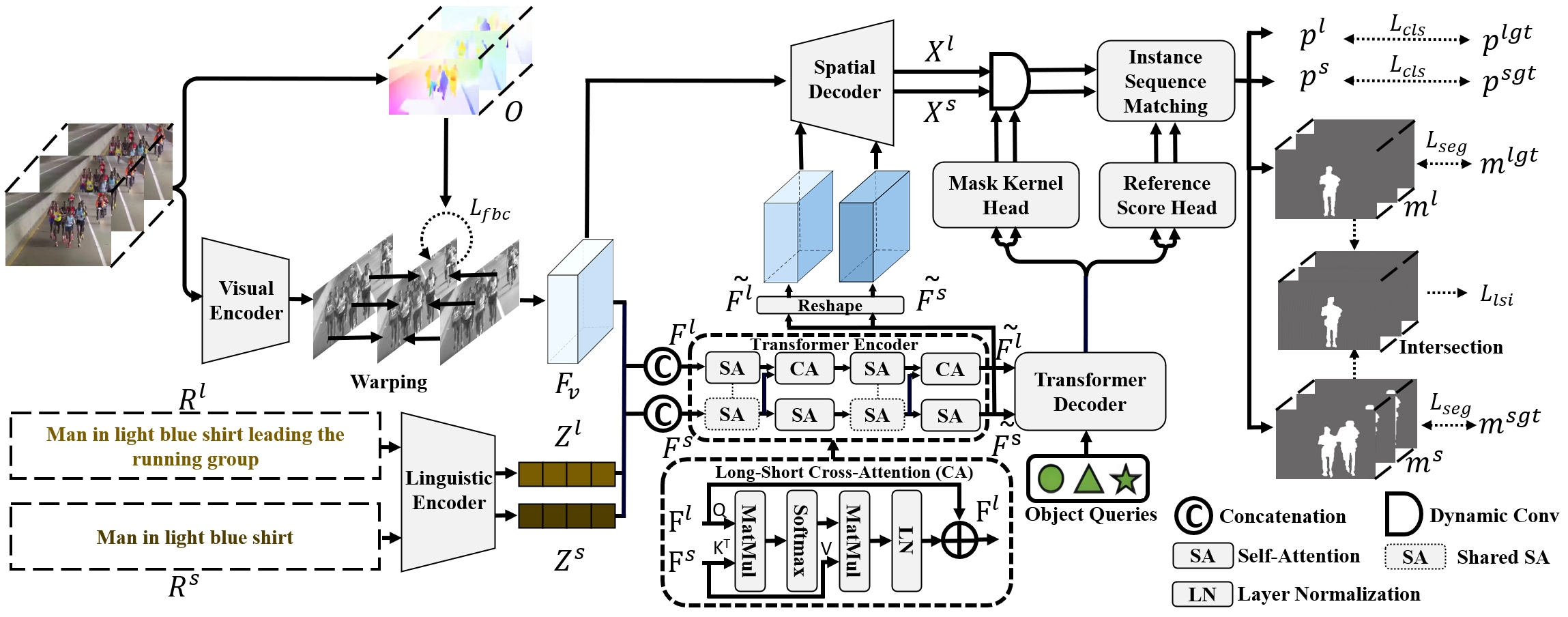}
\end{center}
\setlength{\abovecaptionskip}{-0.3cm}
\setlength{\belowcaptionskip}{-0.3cm}
   \caption{The overall pipeline of LoSh built upon the query-based model~\cite{MTTR}.  Our model takes long and short text expressions as text inputs and uses them to guide the target instance's segmentation in the given video. A long-short cross-attention module, a long-short predictions intersection loss ($\mathcal L_{lsi}$) and a forward-backward visual consistency loss ($\mathcal L_{fbc}$) are specifically introduced. Note that feed-forward networks in transformer encoder are omitted for simplicity.  %Note that ${p}^\text{l}$ and ${p}^\text{s}$ are respectively the probabilities that the predicted instances are referred by $\mathcal R^\text{l}$ and $\mathcal R^\text{s}$. }
   %\miaojing{maybe give some explain on symbols like pl ps and mlgt }
   }
   %Note that these components can be be easily adapted to other methods, like~\cite{ReferFormer, SgMg}.}
\vspace{-6pt}
\label{fig:overview}
\end{figure*}

%generates region proposals for the objects in the video at the first stage. Then the linguistic features of referring sentences are decomposed into actor-/action-related parts.  Independent actor and action modules are utilized to predict the actor-/action-related scores for all region proposals. The proposal with the highest scores is selected and used to generate the mask predictions.

%Among the eaThe single-stage methods cannot obtain significant performance, while the multi-stage methods suffer from heavy workloads because of multi-stage pipelines. 

Recently, owing to the success of query-based transformer models in computer vision~\cite{DETR, DeformableDETR, VisTR,zhou2023hilo}, a simple and unified framework, MTTR~\cite{MTTR}, is firstly introduced in RVOS. It is adapted from the query-based detection work, DETR~\cite{DETR}, where a set of trainable object queries are utilized in the transformer decoder to generate predictions. Furthermore, ReferFormer~\cite{ReferFormer} 
retains the transformer architecture but utilizes a much smaller set of object queries conditioned on text expressions. 
SgMg~\cite{SgMg} uses segmentation optimizer to replace the spatial decoder in ReferFormer~\cite{ReferFormer} and leverages spectrum information to guide the fusion of visual and textural features. 
TempCD~\cite{TempCD} introduces a global referent token and uses collection and distribution mechanisms to interact information between the referent token and object queries. 
OnlineRefer~\cite{OnlineRefer} breaks up the offline belief in previous query-based RVOS frameworks~\cite{MTTR,ReferFormer} and proposes an online RVOS framework using explicit query propagation.
% \textcolor{red}{HTML~\cite{Han2023} introduces an inter-scale vision-language perception module where the linguistic queries dynamically interact with visual features across multiple temporal scales.
% SOC~\cite{Luo2024} facilitates joint space learning by associating a group of frame-level object embeddings with language
% tokens and presenting multi-modal contrastive supervision.} 
{There are other works focusing on temporal modeling based on the query-based transformer framework, \eg,  HTML~\cite{Han2023} 
%interacting multi-modal features on different temporal scales, 
and SOC~\cite{Luo2024}.}
%facilitating joint space learning across modalities and time steps.}
% \textcolor{red}{Therefore, we further exploit the correlation between original expressions and our crafted appearance-centric expressions at both the feature and prediction levels.}

Without loss of generality, we can build our work on top of the recent query-based transformer frameworks given their efficiency and superiority. {Prevailing methods typically utilize complete referring expressions and focus on exploiting the cross-modal feature interactions. Our idea instead utilizes a subject-centric short text expression from the original long expression for joint predictions.} It shares some similarities to~\cite{AAMN, CMPC}, but having a closer look, they are fundamentally different. In~\cite{AAMN}, the idea of predicting the actor- and action-related scores can only work for the two-stage pipeline in which the instance proposals are given at the first stage. It can not be adapted to the single-stage pipeline by directly predicting the masks, as neither the actor- nor action-related words in text expressions are sufficient for the  segmentation of the referred target instance. In~\cite{CMPC}, given the input text expression, it extracts features of individual words and classifies them as \emph{entity}, \emph{relation}, \etc. The entity and relation features are obtained via the weighted combination of word features by using word classification probabilities as weights. However, the word classification in~\cite{CMPC} is not supervised by ground truth word types but is implicitly supervised by the segmentation loss, which can not be fully reliable.

\noindent
\textbf{Transformer.}  Transformer~\cite{Transformer} has a strong ability to draw long-term global dependencies and has achieved huge success in NLP tasks  \cite{GPT3, BERT}. After the introduction of vision transformer~\cite{VIT}, transformer-based models have shown promising results in object detection~\cite{DETR, DeformableDETR}, semantic segmentation~\cite{segmenter, SOTR}, and multi-modal tasks~\cite{CLIP,zhou2023vlprompt}. DETR~\cite{DETR} first introduces the query-based mechanism in object detection. It utilizes a set of object queries as box candidates and predicts the final box embeddings in the transformer decoder. 
%This mechanism simplifies the pipeline and gains significant performance. 
In video instance segmentation, VisTR~\cite{VisTR} employs the idea of DETR and models the task as a sequence prediction problem to perform natural instance tracking. Later on, MTTR~\cite{MTTR}, ReferFormer~\cite{ReferFormer}, SgMg~\cite{SgMg}, \etc extend the DETR~\cite{DETR} and VisTR~\cite{VisTR} into the RVOS task and gain significant improvements. 
%due to the powerful ability of Transformer-based networks to extract and align visual and linguistic features. \miaojing{Later on, MTTR and xxx also xx}. \linfeng{Done.} 
%Inspired by these developments, our method is also built upon the query-based Transformser frameworks and gains significant performance in end-to-end manners.
\section{Method}

\subsection{Problem setting}
The input of RVOS framework consists of a video clip  $\mathcal{V}=\left\{v_t\right\}_{t=1}^T$, $v_t \in \mathbb{R}^{3 \times H \times W}$ with \textit{T} frames and a text expression $\mathcal{R}=\left\{r_l\right\}_{l=1}^L$ with \textit{L} words. The aim of this task is to generate pixel-wise mask predictions $\mathcal{M}=\left\{m_t\right\}_{t=1}^T$, $m_t \in \mathbb{R}^{{H}' \times {W'}}$ for the target instance referred by $\mathcal{R}$ in several frames of $\mathcal{V}$.

\subsection{Preliminary}
\label{preliminaries}
%\noindent
%\textbf{Task definition.} 

%The basic idea of LoSh is pluggable on 
As shown in Fig.~\ref{fig:overview}, we build the proposed LoSh upon state of the art query-based transformer model, MTTR~\cite{MTTR}.  
%and SgMg~\cite{SgMg}, noting that MTTR and SgMg share a similar basic architecture. 
Without loss of generality,  LoSh can also be plugged into many other RVOS baselines.  %and compatible with all the RVOS frameworks. Since the query-based model is not the focus of our approach, 
Below, we briefly review this query-based transformer framework for RVOS.

\noindent
\textbf{Visual encoder.} Given the input frames in $\mathcal{V}$, visual features are extracted via a 
%The query-based methods extract the visual features from  by a spatial-temporal backbone.
spatial-temporal visual encoder, namely the Video Swin transformer~\cite{videoswint}, such that   $\mathcal{F}_v=\left\{f_t\right\}_{t=1}^T$,  $f_t \in \mathbb{R}^{c_1 \times h \times w}$ corresponding to the $t$-th frame in $\mathcal V$. This visual encoder 
%extends the original Swin Transformer\cite{swint} with a temporal down-sampling layer \miaojing{only this difference?} and 
is pre-trained on Kinetics-400~\cite{kinetics400}, which can simultaneously aggregate spatial and temporal information in a video. 

\noindent
\textbf{Linguistic encoder.} Given the input words in $\mathcal R$, linguistic features are extracted using the linguistic encoder, RoBERTa~\cite{roberta}, such that $\mathcal{Z} = \left\{z_l\right\}_{l=1}^L$,  $z_l \in \mathbb{R}^{c_2}$ corresponding to the $l$-th word in $\mathcal R$. This linguistic encoder 
%extends the original Swin Transformer\cite{swint} with a temporal down-sampling layer \miaojing{only this difference?} and 
is pre-trained on several English-language corpora~\cite{LanguageDataset1, LanguageDataset2}.

\noindent
\textbf{Mask generation.}  
%At the same time, linguistic features are extracted from the language expression $\mathcal{T}$ by a Transformer-based text encoder. 
Having $\mathcal{F}_v$ and $\mathcal Z$, 
%in order to align them,
their channel dimensions are scaled to $\mathcal{D}$ (\eg, 256) by a $1\times1$ convolutional layer and a fully connected layer, respectively. 
%(\miaojing{why not use both 1 times 1 conv or fc?} \linfeng{The 1x1 conv and fc play the same role here. But the F are feature maps while the Z are vectors.} ) 
Then, they are fused in the transformer encoder, which is devised on top of visual and linguistic encoders, to generate the multi-modal features, $\widetilde{\mathcal F}$.

In the transformer decoder, a set of trainable object queries are utilized to predict entity-related information. Each object query corresponds to a potential instance. The same query across multiple frames is trained to represent the same instance in the video. This design allows the natural tracking of each instance in the video. Supposing that the number of object queries is $N$, we can get $N$ instance sequences, where the length of each sequence is $T$. On top of the decoded object queries, several lightweight heads are utilized to predict the reference scores and segmentation kernels. 
%, or \miaojing{and?} \linfeng{Here we use 'or' because only ReferFormer predicts boxes for object instances while MTTR doesn't do so.} bounding boxes. 
The reference score indicates the probability that the predicted instance is the referred target instance. 
%\linfeng{Describe the reference score here to answer the previous question A.} 
The predicted segmentation kernels,  inspired by other works in instance segmentation~\cite{kernel1, kernel2}, are used to convolve the text-related video features, $\mathcal X$. $\mathcal X$ is generated via a cross-modal FPN-like  spatial decoder which takes the input of  $\mathcal{F}_v$ and $\widetilde{\mathcal F}$, 
%($\mathcal F$ after transformer encoder), 
as illustrated in Fig.~\ref{fig:overview}.
%$ are input to cross-modal FPN. The text-related video features mentioned here mean the output of the cross-modal FPN.} by a cross-modal FPN-like\cite{fpn} module which takes $\mathcal F$ and $\mathcal X$ as inputs. 
%\linfeng{I believe this process can be much more clear after completing the Figure of the overall review of LoSh.}
%\linfeng{Answer question B. It also illustrates the position of the FPN-like module.} 
For more details of this generation process, we refer the readers to \cite{MTTR}.

\subsection{Long-short text expressions}
\label{sec:long-short text expression}
The long text expression ($\mathcal R^\text{l}$) for each video clip is given in the training set (\eg, `a man in a white t-shirt is walking'), which normally consists of a subject, predicate, object, some adjectives and descriptive phrases describing the subject or the object. We can generate the subject-centric short text expression ($\mathcal R^\text{s}$) by removing the predicate- and object-related contents in the long text expression. This can be either manually done or automatically achieved via a part-of-speech tagging method provided by \cite{NLTK}.
%The proposed LoSh directly extracts the subject and the following description of its appearance as short input text by hand or part-of-speech tagging method\cite{POStagging}. 
For the latter,  we can let the part-of-speech tagger identify the position of the first verb in a text expression and keep the words before it to create a subject-centric short text expression.

\subsection{Long-short mask predictions}
\label{sec:long-short encoder blocks}
%\textcolor{red}{We use two branches (\ie, \textit{long-text} and \textit{short-text}) to generate predictions for $\mathcal R^\text{l}$ and $\mathcal R^\text{s}$:}
$\mathcal R^\text{l}$ and $\mathcal R^\text{s}$ are fed into the linguistic encoder simultaneously, resulting in text embeddings $\mathcal{Z}^\text{l}$ and $\mathcal{Z}^\text{s}$. 
% Following what is outlined in Sec.~\ref{preliminaries}
As illustrated in Fig.~\ref{fig:overview}, they are respectively concatenated with the same visual features $\mathcal{F}_v$ to generate  $\mathcal F^\text{l}$ and $\mathcal F^\text{s}$. 
% Then, $\mathcal F^\text{l}$ and $\mathcal F^\text{s}$ are fed into our proposed simple-yet-effective long-short (LoSh) encoder blocks.
After the transformer encoder, we obtain multi-modal features $\widetilde{\mathcal F^\text{l}}$ and $\widetilde{\mathcal F^\text{s}}$, we interact each object query with them in the transformer decoder to obtain the soft mask predictions for the referred instance by $\mathcal R^\text{l}$ and $\mathcal R^\text{s}$, respectively. Sharing queries between $\widetilde{\mathcal F^\text{l}}$ and $\widetilde{\mathcal F^\text{s}}$ would facilitate the model training.
%Specifically, we duplicate the trainable object queries and let them interact with $\mathcal F^\text{l}$ and $\mathcal F^\text{s}$ in the decoder respectively to generate predictions for $\mathcal R^\text{l}$ and $\mathcal R^\text{s}$. By utilizing the duplicated object queries, we can avoid introducing and learning new trainable object queries, thus facilitating the network training.
% By utilizing duplicated object queries, we can generate predictions for $\mathcal R^\text{l}$ and $\mathcal R^\text{s}$ simultaneously and 
% avoid learning additional object queries, thus facilitating model training.
%Through this approach, each object query can generate predictions for $\mathcal R^\text{l}$ and $\mathcal R^\text{s}$ simultaneously, reducing the number of required object queries and facilitating the network training.
% }
The generated soft masks are denoted by ${\mathcal{M}}^\text{l}=\left\{m^\text{l}_t\right\}_{t=1}^T$ and  ${\mathcal{M}}^\text{s}=\left\{m^\text{s}_t\right\}_{t=1}^T$,
%$\hat{\mathcal{S}}_{short}=\left\{s_{short}^t\right\}_{t=1}^T$, $s_{short} \in \mathbb{R}^{H \times W}$ 
%corresponding to each object query
where $m$ signifies a probability map after the sigmoid operation, each of its pixel value indicating the probability that this pixel belongs to the referred instance. 
Since $\mathcal R^\text{l}$ is a sophisticated description of the target instance, we observe ${\mathcal{M}}^\text{l}$ tends to favour more on the instance’s action rather than appearance (see Fig.~\ref{fig:long} and Fig.~\ref{fig:qualitative_comparison}). To strengthen the appearance-related information in $\mathcal F^\text{l}$, we introduce a long-short cross-attention module within the transformer encoder to take advantage of the subject-centric appearance information encoded in $\mathcal F^\text{s}$. 
%
% exclusively encoded in $\mathcal R^\text{s}$.  

%Given long and short language expressions respectively, 
%\linfeng{clarify the meaning of M, a pixel-wise probability for the foreground.}
%We denote by the predictions for long and short text as $\hat{\mathcal{S}}_{long}=\left\{s_{long}^t\right\}_{t=1}^T$, $s_{long} \in \mathbb{R}^{H \times W}$ and $\hat{\mathcal{S}}_{short}=\left\{s_{short}^t\right\}_{t=1}^T$, $s_{short} \in \mathbb{R}^{H \times W}$.  

% \begin{figure}
% \begin{center}

% \includegraphics[width=0.9\linewidth]
% {losh_encoder.png}

% \end{center}
%    \caption{\textcolor{red}{Illustration of our proposed long-short cross-attention mechanism. Each LoSh encoder block (black dotted frame) consists of a self-attention encoding block (\textcolor{myorange}{orange} dotted frame) and a short-to-long (Sh2Lo) attention encoding block (\textcolor{blue}{blue} dotted frame).}}
%    \vspace{-5pt}
% \label{fig:long-short encoder blocks}
% \end{figure}

\noindent \textbf{Long-short cross-attention.} 
The original transformer encoder performs a series of self-attention (SA) within $\mathcal F^\text{l}$ or $\mathcal F^\text{s}$ separately. 
%{In order to facilitate the interaction between $\mathcal F^\text{l}$ and $\mathcal F^\text{s}$, 
%we introduce a long-short cross-attention (CA) module. 
%In the transformer encoder of LoSh, $\mathcal {F}^\text{l}$ is strengthened by $\mathcal{F}^\text{s}$ through alternating SA and CA while  $\mathcal{F}^\text{s}$ is treated by standard SA.} 
In order to facilitate the interaction between $\mathcal F^\text{l}$ and $\mathcal F^\text{s}$, we inject a long-short cross-attention (CA) module to replace the even-numbered self-attention module of the transformer encoder.
Specifically, the proposed long-short cross-attention module treats $\mathcal F^\text{s}$ as \textit{key} and \textit{value} while $\mathcal F^\text{l}$ as \textit{query}. In this way, $\mathcal F^\text{l}$ can be strengthened by aggregating appearance-related information in $\mathcal F^\text{s}$, therefore alleviating the RVOS model's excessive focus on the action-related information of the target instance. We design the cross-attention to be uni-directional between $\mathcal F^\text{s}$ and $\mathcal F^\text{l}$ because our ultimate target is to utilize the auxiliary information in $\mathcal F^\text{s}$ to help accurately segment the instance referred by $\mathcal R^\text{l}$ ($\mathcal F^\text{l}$). Empirically, we also did not observe benefits by using $\mathcal F^\text{l}$ to strengthen $\mathcal F^\text{s}$ (see Sec.~\ref{ablation study}). We suggest the reason is this can conversely down-weight the appearance-related information in $\mathcal F^\text{s}$, which is the key information we care about in $\mathcal F^\text{s}$. 

\medskip 
Overall, the new transformer encoder now has two pathways (see Fig.~\ref{fig:overview}), one for $\mathcal F^\text{s}$ containing only SA modules same to the original version, one for $\mathcal F^\text{l}$ alternating between SA and CA modules. 
%The enhanced $\mathcal F^\text{l}$ is aggregated from $\mathcal F^\text{s}$ through the attention map between \textit{query} and \textit{key}, and $\mathcal F^\text{s}$ only focuses on the appearance-related multi-modal information of the target instance.
%This unidirectional mechanism only utilizes $\mathcal F^\text{s}$ to enhance $\mathcal F^\text{l}$, but does not use $\mathcal F^\text{l}$ to strengthen $\mathcal F^\text{s}$. 

%Functionally, the MHSA is responsible for capturing global information of $\mathcal F^\text{s}$. 

%Given the same input video frames, $\mathcal R^\text{s}$ is the subject-centric part of $\mathcal R^\text{s}$, making $\mathcal R^\text{s}$ a more generic expression of a certain subject than $\mathcal R^\text{l}$. Therefore, the multi-modal feature $\mathcal F^\text{s}$ corresponding to $\mathcal R^\text{s}$  focuses on the subject with a certain appearance while $\mathcal F^\text{l}$ corresponding to $\mathcal R^\text{l}$ pays attention to the subject with both a certain appearance and action, though the action is often more favoured.

%Due to the different granularity of the input $\mathcal R^\text{l}$ and $\mathcal R^\text{s}$, the $\mathcal F^\text{l}$ is more specific than $\mathcal F^\text{s}$ which exclusively focuses on the appearance-related multi-modal information of the target instance. An intuitive idea is that through the proper feature interaction between $\mathcal F^\text{l}$ and $\mathcal F^\text{s}$, more robust multi-modal features and mask predictions can be obtained. 

\subsection{Long-short predictions intersection}
\label{sec:long-short intersection loss}
Given the same input video frames,  the short text expression is the subject-centric part of the long one, making it a more generic expression of a certain subject than the longer one. 
%The mask prediction for the short text expression focuses on the subject with a certain appearance while that for the long text expression pays attention to the subject with both a certain appearance and action, though the action is often more favoured. 
They focus on different levels of details and  the mask prediction for the short text expression should be ideally included in that for the long one. To regulate our network predictions to conform to this observation, we introduce a new long-short predictions intersection loss, specified below.    

%We introduce a novel conditional IOU into the matching cost and loss functions in query-based RVOS frameworks. 

% Ideally, $\left|{{m}}_t^\text{l} \cap {m}_t^\text{s}\right|$ should be equivalent to $\left|{m}_t^\text{l}\right|$,  \ie, $\frac{\left|{m}_t^\text{l} \cap  {m}_t^\text{s}\right|}{\left|m_t^\text{l}\right|}  = 1$.

Given ${m}_t^\text{l}$ and ${m}_t^\text{s}$, we first use a threshold (\eg, 0.5) to filter out those pixels whose values are below it. We consider they are more likely to be the background pixels.
%\miaojing{probabilities it should have high values for the bg class} \linfeng{We have clarified that mt represents the foreground probability of pixels.} 
Next, we compute the intersection between ${{m}}_t^\text{l}$ and ${m}_t^\text{s}$ by pixel-wisely multiplying them (after the background removal) and summing their products to obtain the probability weighted intersection area between them, \ie, $\left|{{m}}_t^\text{l} \cap {m}_t^\text{s}\right|$.  We also calculate the sum of the probabilities of foreground pixels in ${m}_t^\text{l}$, \ie, $\left|{m}_t^\text{l}\right|$.  Theoretically, the mask prediction by the long text expression should be included in that by the short text expression; in other words, $\left|{{m}}_t^\text{l} \cap {m}_t^\text{s}\right|$ should be equivalent or at least very close to $\left|{m}_t^\text{l}\right|$. 
In practice, 
$\frac{\left|{m}_t^\text{l} \cap  {m}_t^\text{s}\right|}{\left|m_t^\text{l}\right|}$ belongs to  $[0, 1]$.  In network optimization, we maximize $\frac{\left|{m}_t^\text{l} \cap  {m}_t^\text{s}\right|}{\left|m_t^\text{l}\right|}$ to encourage the agreement between the long-short predictions on the referred instance; { in other words, enforcing a partial alignment between them.}
The long-short predictions intersection loss is written as:

\vspace{-5pt}
\begin{equation}
\label{conditional_IOU}
\mathcal{L}_{lsi} = \sum_{t=1}^T \left(1- \frac{{\left|{m}_t^\text{l} \cap  {m}_t^\text{s}\right|} +\epsilon}{{\left|m_t^\text{l}\right|}+\epsilon} \right)
\end{equation}

%We regard this calculated intersection as the numerator. The denominator of conditional IOU is the sum of the values of foreground pixels in $\hat{\mathcal{S}}_{long}$. 
We add a constant $\epsilon$ (1.0) to Eqn.~\ref{conditional_IOU} in case $\left|m_t^\text{l}\right|$ is zero. 
%\miaojing{you have used M as either the area or the probabilities, this is wrong} \linfeng{The M is regarded as probabilities in Method Section all the time.}
$\mathcal{L}_{lsi}$ is integrated into the matching cost (below) to find the best-matched object query for the ground truth. {For cases when ${m}_t^\text{l}$ and ${m}_t^\text{s}$ are completely disjoint during training, meaning at least one of them is mis-predicted, they will not only be penalized by $\mathcal{L}_{lsi}$, the mis-predicted one will also be penalized by its own segmentation loss (see Eqn.~\ref{matching cost}).}
%this can eventually lead to better aligned predictions.} 
%\textcolor{blue}{Note that the short text expression may correspond to multiple instances (see Sec.~\ref{implementation}). In these cases, the proposed $\mathcal{L}_{lsi}$ still enables the optimization to focus on the pixels belonging to the predicted target instance $\left(m_t^\text{l}\right)$. }

%\miaojing{I removed something that should be reverted, about this loss will be integrated into the matching cost} \linfeng{Done.}
%\noindent
%This IOU represents the degree of overlap between $\hat{\mathcal{S}}_{long}$ and $\hat{\mathcal{S}}_{short}$ within the foreground pixels in $\hat{\mathcal{S}}_{long}$. $\hat{\mathcal{S}}_{short}$ is closer to the basic attributes (\eg, shape, color, appearance) while $\hat{\mathcal{S}}_{long}$ focus on complex characteristics(\eg, appearance, actions, relations). Ideally,  $\hat{\mathcal{S}}_{long}$ is subset of  $\hat{\mathcal{S}}_{short}$. 
%During training, $\hat{\mathcal{S}}_{long}$ and $\hat{\mathcal{S}}_{short}$ have overlapping and disjoint parts. The predicted pixels in the overlaps are the masks for both appearance-focused short sentences and action-focused long sentences. They are reliable and should be encouraged to predict with higher confidence. In contrast, the rest predicted pixels in $\hat{\mathcal{S}}_{long}$ correspond to un-referred objects. These predictions contradict the appearance-focused $\hat{\mathcal{S}}_{long}$ and should be penalized. 

\subsection{Long-short matching cost}
According to the mask prediction process in  Sec.~\ref{preliminaries}, the query-based RVOS framework generates $T \times N$ predictions where \textit{T} is the number of input frames and \textit{N} is the number of object queries. The predictions of each object query %\miaojing{which prediction} \linfeng{Here emphasize the predictions of object queries} 
maintain the same relative positions across different frames. 
%Therefore, the tracking process can be performed naturally. 
The predictions for the $i$-th object query can be denoted by:

\vspace{-5pt}
\begin{equation}
{y_i} = \left\{ \left({p}_{i,t}^\text{l}, {m}_{i,t}^\text{l}\right), \left({p}_{i,t}^\text{s}, {m}_{i,t}^\text{s}\right) \right\}_{t=1}^T
\end{equation}
\vspace{-7pt}

\noindent
where ${p}_{i,t}^\text{l}$ (${p}_{i,t}^\text{s}$) is the probability that the $i$-th object query corresponds to the instance referred by $\mathcal R^\text{l}$ ($\mathcal R^\text{s}$) in the $t$-th frame.
${m}^\text{l}_{i,t}$ (${m}^\text{s}_{i,t}$) is the corresponding mask prediction by the $i$-th query in the $t$-th frame given $\mathcal R^\text{l}$ ($\mathcal R^\text{s}$).

In contrast, the ground truth is represented by:

\vspace{-5pt}
\begin{equation}
y^{gt} = \left\{ \left(p_t^{lgt},  m_t^{lgt}\right), \left(p_t^{sgt}, m_t^{sgt}\right)\right\}_{t=1}^T
\end{equation}
\vspace{-5pt}

\noindent
where $p_t^{lgt}$ ($p_t^{sgt}$) equals to $1$ 
when the {referred} instance is visible in the $t$-th frame, otherwise 0. $m_t^{lgt}$ ($m_t^{sgt}$) is the ground truth mask for $\mathcal R^\text{l}$ ($\mathcal R^\text{s}$) in the $t$-th frame. In most cases, {$m_t^{lgt}$ and $m_t^{sgt}$ are the same corresponding to one specific instance (so as $p_t^{lgt}$ and $p_t^{sgt}$), though there exist cases when $m_t^{sgt}$ corresponds to multiple instances (see also Sec.~\ref{implementation}). We can handle both situations. 

%of the referred target corresponding to language expressions. \miaojing{pt and mt is xxx.} \linfeng{Done.}

Given $y^{gt}$, the best-matched prediction ${y^*}$ over $\mathcal Y = \{y_i\}_{i=1}^N$ is found by minimizing a long-short matching cost:

\vspace{-5pt}
\begin{equation}
 {y^*} = \underset{{y}_i \in \mathcal{Y}} {{\arg\min} \,} \mathcal{L}_{lsm}\left({y_i}, y^{gt}\right)
\end{equation}
\vspace{-5pt}

\noindent 
Based on the matching cost in query-based RVOS frameworks~\cite{MTTR, SgMg}, we develop the long-short matching cost as: 
\vspace{-5pt}
\begin{equation}
\label{matching cost}
\begin{split}
\mathcal{L}_{lsm}\left({y}_i, y^{gt}\right) = 
\sum_{t=1}^{T}
\Bigr [
\lambda_{lsi}\mathcal{L}_{lsi}\left(m_{i,t}^\text{l}, m_{i,t}^\text{s}\right)\\ +
\lambda_{cls}\left(\mathcal{L}_{cls}\left(p_{i,t}^\text{l}, p_t^{lgt}\right) +
\mathcal{L}_{cls}\left(p_{i,t}^\text{s}, p_t^{sgt}\right)\right) \\ +
\lambda_{seg}\left(\mathcal{L}_{seg}\left(m_{i,t}^\text{l}, m_t^{lgt}\right)   + \mathcal{L}_{seg}\left(m_{i,t}^\text{s}, m_t^{sgt}\right) \right)
\Bigl ]
\end{split}
\end{equation}
\vspace{-5pt}

\noindent
{where $\mathcal{L}_{lsi}$ is the long-short predictions intersection loss introduced in Eqn.~\ref{conditional_IOU}. $\mathcal{L}_{cls}$ signifies the  binary classification loss while $\mathcal{L}_{seg}$ the segmentation loss.} Each segmentation loss consists of a DICE loss and a mask focal loss according to MTTR~\cite{MTTR}.  $\lambda_{lsi}$, $\lambda_{seg}$, and $\lambda_{cls}$  are hyper-parameters.

\subsection{Forward-backward visual consistency}
\label{sec:forward-backward consistency}
%So far, we have focused on 
{Previously, we focus on} the linguistic aspect of the RVOS by employing long-short text expressions to exploit  the spatial correlations between pixels (segments) within each frame.
In this section, we focus on the visual aspect of the RVOS by employing the optical flow to exploit the temporal correlations between pixels across frames. 
%propose an improvement in the visual aspect, which can strengthen the temporal consistency of the pixels in feature maps across multiple frames. \linfeng{Emphasize the relationship between the long-short intersection and past-future consistency. language vs vision. spatial vs temporal.} To achieve this goal, we introduce a temporal-related visual cue, optical flow. 
Optical flow represents the motion of pixels between consecutive video frames, caused by the movement of instances or the camera. 
It plays an important role in video segmentation \cite{EFC, videosegviaflow, videosegbyrecurrentflow} and action recognition~\cite{actionrecognitionintegrationflow, actionrecognitionmotionrepresentation}.  Particularly, \cite{EFC} devises a learnable module to estimate optical flow between two video frames and utilize it for  occlusion modeling and feature warping. Inspired by it, and since there are nonconsecutive annotated frames in each video clip, we employ forward and backward optical flows to warp every annotated frame's neighbors to it for visual consistency.    
Specifically, we first calculate the  optical flows from a certain annotated frame (\eg, $k$-th frame) to their adjacent four frames,
%\miaojing{xx and its adjacent frames xxx} \linfeng{Done.}, 
which are denoted by $\mathcal{O}=\left\{o_{k \rightarrow k+t}\right\}_{t=-2}^{2}$, $o_{k \rightarrow k+t} \in \mathbb{R}^{H \times W}$. Unlike~\cite{EFC}, we use the Farneback method ~\cite{DenseOpticalFlowEstimation} to directly compute $\mathcal{O}$ without learning given the video frames. The forward optical flows, $\mathcal{O}_f=\left\{o_{k \rightarrow k+t}\right\}_{t=1}^{2}$, are utilized to warp the visual features of $\left(k+t\right)$-th frames to the $k$-th frame; the backward optical flows, $\mathcal{O}_b=\left\{o_{k \rightarrow k-t}\right\}_{t=1}^{2}$ are utilized to  warp the visual features of $\left(k-t\right)$-th frames to the $k$-th frame.
%\miaojing{We can use \miaojing{xx} to forwardly warp the xx frame to the xx frame; and xxx to backwardly warp the xx frame to the xx frame.}\linfeng{Done.}  
We denote the warped features of $k$-th frame by $\mathcal{F}^w = \left\{f_{k+t \rightarrow k}^{w}\right\}_{t=-2}^{2} $ ($f^w_{k \rightarrow k}$ is equivalent to $f_k$). 
%\linfeng{Replace the equation containing 'Warp' only.}
%exploit the calculated optical flows to compute the warped features of the keyframe:
% \begin{equation}
% f_{i \rightarrow k}^{warped}=Warp\left(f_i, o_{k \rightarrow i}\right)
% \end{equation}
% \noindent
The warped features and the original visual feature of the $k$-th frame should be similar, we minimize their distances and write out our forward-backward visual consistency loss as:
%The warped feature we calculate only the warped features of the two frames before and after the keyframe because the baseline models take 5 to 8 frames as input.Then, $L_2$ norm is utilized to calculate the distances between the keyframe features and warped ones:
\vspace{-5pt}
\begin{equation}
\label{fbc}
\mathcal{L}_{fbc}\left(\mathcal F^{w}, f_k\right)  = \sum_{t=-2}^{2}\Vert f_{k+t \rightarrow k}^{w} - f_k \Vert_2
\end{equation}
\vspace{-7pt}

%Through this past-future consistency loss, the cross-frame feature consistency is enhanced. 
This loss enhances the representation of  semantic- and motion-related information in the visual features $\mathcal{F}_v$. 
%effectively improves the temporal information extraction ability of the visual backbone for RVOS. 

\subsection{Network training and inference}
\label{training_and_inference}
\noindent \textbf{Network training.}
After getting the best-matched prediction ${y}^*$, the final loss function is a combination of the matching cost in Eqn.~\ref{matching cost} and the forward-backward consistency loss in Eqn.~\ref{fbc} with a hyperparameter $\lambda_{fbc}$:
\begin{equation}
\label{lossfunction}
\begin{split}
\mathcal{L}\left({y}^*, y^{gt}\right) = 
 \mathcal{L}_{lsm}\left({y}^*, y^{gt}\right) + \lambda_{fbc}\mathcal{L}_{fbc}\left(\mathcal F^w, f_k\right)
\end{split}
\end{equation}

% \begin{equation}
% \label{lossfunction}
% \begin{split}
% \mathcal{L}\left({y}_b, y^{gt}\right) = 
% \sum_{t=1}^{T}
% \Bigr [
% \lambda_{cls}\mathcal{L}_{cls}\left(p_{b,t}^\text{l}, p_t^{lgt}\right)\\ +
% \lambda_{cls}\mathcal{L}_{cls}\left(p_{b,t}^\text{s}, p_t^{sgt}\right)
% + \lambda_{seg}\mathcal{L}_{seg}\left(m_{b,t}^\text{l}, m_t^{lgt}\right)   \\  +
% \lambda_{seg}\mathcal{L}_{seg}\left(m_{b,t}^\text{s}, m_t^{sgt}\right) 
% + \lambda_{lsi}\mathcal{L}_{lsi}\left(m_{b,t}^\text{l}, m_{b,t}^\text{s}\right)
% \Bigl ]  \\ + \lambda_{fbc}\mathcal{L}_{fbc}\left(\mathcal F^w, f_k\right)
% \end{split}
% \end{equation}

\noindent \textbf{Network inference.} 
We deactivate the prediction head for the short text expression but use the long text expression for inference. 
Following Sec.~\ref{preliminaries},
%we input long-short text expressions together into the model. After the transformer encoder, the short text computation branch is discarded to ensure computational efficiency. Besides, optical flow does not need to be considered during training. Therefore, our network runs only slightly slower than the baseline method.} As in
LoSh generates $N$ instance sequences, 
%where the length of each sequence is $T$. 
each containing reference scores and mask predictions for the predicted instance in $T$ frames.  
%\linfeng{Put the detailed description in Section 3.2 and refer to the section.} 
We average the reference scores of predicted instance across frames. The sequence with the highest average reference score is regarded as the final prediction. The final mask predictions are obtained from the selected sequence. 
\section{Experiments}

\begin{table*}
\centering
\begin{tabular}{c|c|ccc|ccc|ccc}
\hline
\multirow{2}{*}{Method} & \multirow{2}{*}{\centering Backbone} & \multicolumn{3}{c|}{A2D-Sentences} & \multicolumn{3}{c|}{Refer-YouTube-VOS} & \multicolumn{3}{c}{Refer-DAVIS17} \\
\cline{3-11}
 &  & O-IoU & M-IoU & mAP & $\mathcal{J} \& \mathcal{F}$ & $\mathcal{J}$ & $\mathcal{F}$ & $\mathcal{J} \& \mathcal{F}$ & $\mathcal{J}$ & $\mathcal{F}$ \\
\hline\hline
Gavrilyuk \etal \cite{firstRVOS} &  I3D & 53.6 & 42.1 & 19.8 & - & - & - & - & - & - \\
CMPC-V~\cite{CMPC} & I3D & 65.3 & 57.3 & 40.4 & 47.5 & 45.6 & 49.3 & - & - & - \\
YOFO~\cite{YOFO} & ResNet-50 &- & - & - & 48.6 & 47.5 &49.7 & 53.3 &48.8 &57.8 \\
MLRL~\cite{MLRL} & ResNet-50 & - & - & -  & 49.7& 48.4 & 51.0 & 52.7 & 50.1 & 55.4 \\
% ReferFormer~\cite{ReferFormer} & Video-Swin-T & 77.6 & 69.6 & 52.8 & 59.4 &58.0& 60.9 & 59.4 &58.0& 60.9 \\
ReferFormer~\cite{ReferFormer} & Video-Swin-B & 78.6 & 70.3 & 55.0 & 62.9 &61.3& 64.6 &61.1 &58.1 &64.1 \\

\hline

MTTR ${\left(w = 8\right)}$~\cite{MTTR} & Video-Swin-T & 70.2 & 61.8 & 44.7 & - & - & - & - & - & - \\
\rowcolor{mygray}
\textbf{LoSh-M ${\left(w = 8\right)}$} & Video-Swin-T & \textbf{72.9}& \textbf{64.9} & \textbf{47.8} & - & - & - & - & - & -\\
\hline
SgMg $\left(w=5\right)$~\cite{SgMg} & Video-Swin-T&  78.0& 70.4& 56.1& 62.0& 60.4& 63.5 & 61.9& 59.0& 64.8 \\
\rowcolor{mygray}
\textbf{LoSh-S $\left(w=5\right)$} & Video-Swin-T& \textbf{79.3}& \textbf{71.6}& \textbf{57.6}& \textbf{63.7}& \textbf{62.0}& \textbf{65.4} & \textbf{62.9}& \textbf{60.1}& \textbf{65.7}  \\
SgMg $\left(w=5\right)$~\cite{SgMg} & Video-Swin-B&  79.9& 72.0& 58.5& 65.7& 63.9& 67.4 & 63.3& 60.6& 66.0 \\
\rowcolor{mygray}
\textbf{LoSh-S $\left(w=5\right)$} & Video-Swin-B&  \textbf{81.2}& \textbf{73.1}& \textbf{59.9}& \textbf{67.2} &\textbf{65.4} &\textbf{69.0} & \textbf{64.3} &\textbf{61.8} &\textbf{66.8} \\
\hline
\end{tabular}
\caption{Quantitative comparison with state of the art on A2D-Sentences~\cite{firstRVOS}, Refer-YouTube-VOS\cite{Urvos} and Refer-DAVIS17~\cite{refdavis17}. O-IoU and M-IoU represent Overall IoU and Mean IoU. The number of input frames $w$ follows the implementation details of~\cite{MTTR,SgMg}.
%\miaojing{YOU MENTIONED SEVERAL DATASETS IN THE INTRODUCTION.} \linfeng{Done. Add Refer-YouTube-VOS here.}}
}
\vspace{-6pt}
\label{tab:SOTAcomparison_combined}
\end{table*}

\subsection{Datasets and evaluation metrics}
We conduct experiments on four popular RVOS benchmarks: A2D-Sentences~\cite{firstRVOS}, Refer-YouTube-VOS~\cite{Urvos}, JHMDB-Sentences~\cite{firstRVOS} and Refer-DAVIS17~\cite{refdavis17}. A2D-Sentences contains 3,754 videos, with 3,017 for training, and 737 for testing. 
Each video has three annotated frames with pixel-wise segmentation masks for different target instances. It has 6,655 text expressions and each text expression corresponds to only one instance in the annotated frames of one video.
% move to supp
% JHMDB-Sentences has 928 videos with 928 corresponding text expressions in total. The whole data is normally used for evaluation after the RVOS model is trained  on A2D-Sentences~\cite{firstRVOS}.
Refer-YouTube-VOS~\cite{Urvos} is the largest RVOS dataset, containing 3,978 videos with 15,009 text expressions. Each video in this dataset is annotated with pixel-wise instance segmentation masks for every fifth frame. We train our model on the training partition and obtain the results on the validation partition of Refer-YouTube-VOS. 
% remove redundant information about ref-youtube-vos.
% Following~\cite{ReferFormer, SgMg}, we use the subset with the \textit{full-video expressions} to train and evaluate our model. This subset has two partitions:  the first partition includes 12,913 text expressions for 3,471 videos and is used for training. The second partition includes 2,096 expressions for 507 videos and is further divided into 202 and 305 videos for validation and testing respectively. We train our model on the training partition and obtain the results on the validation partition. 
Furthermore, following~\cite{ReferFormer, SgMg}, the models trained on A2D-Sentences and Refer-YouTube-VOS are respectively evaluated on JHMDB-Sentences and Refer-DAVIS17. JHMDB-Sentences and Refer-DAVIS17 are extensions of JHMDB~\cite{visualjhmdb} and DAVIS17~\cite{davis17} with additional text expressions. JHMDB-Sentences has 928 videos with 928 corresponding text expressions while Refer-DAVIS17 contains 90 videos with 1,544 expression sentences in total. 
% Only 30 video clips in the validation split are utilized to evaluate the generalizability of the model trained on Refer-YouTube-VOS.
% Following~\cite{ReferFormer, SgMg}, we also use the subset of Refer-YouTube-VOS with the \textit{full-video expressions} to train and evaluate our model. The trained model from Refer-YouTube-VOS is further evaluated on Refer-DAVIS17 which contains 30 video clips in the validation split.
% More details about Refer-YouTube-VOS and Refer-DAVIS17 can be found in the supplementary material. 
%\linfeng{Add introduction of Refer-YouTube-VOS here.} \miaojing{what about refdavis dataset} \linfeng{Done.}

Following~\cite{firstRVOS, MTTR}, we adopt  Overall IoU, Mean IoU and mAP to evaluate our model on A2D-Sentences and JHMDB-Sentences.  Overall IoU calculates the ratio of the intersection and union of all predicted and annotated segment pixels in the test set. Mean IoU calculates the average of IOU results for each instance over all frames in the validation set. mAP computes the average of mAPs calculated under 10 IoU thresholds between 0.5 and 0.95 over testing samples. For Refer-YouTube-VOS and Refer-DAVIS17, we follow~\cite{ReferFormer} to use region similarity ($\mathcal{J}$),  contour accuracy ($\mathcal{F}$)  and the average of them ($\mathcal{J} \& \mathcal{F}$). The calculation of $\mathcal{J}$ is the same as IoU between prediction and annotation. $\mathcal{F}$ is the F1-score calculated by the precision and recall of the  contour of the predicted mask for the target instance. %\linfeng{Introduce the metrics of Refer-YouTube-VOS here.}

\subsection{Implementation details}
\label{implementation}
{We build the proposed LoSh upon two state of the art pipelines, MTTR~\cite{MTTR} and SgMg~\cite{SgMg} to obtain LoSh-M and LoSh-S, respectively.}

\noindent
\textbf{LoSh-M.} 
% Following MTTR~\cite{MTTR}, 
% we use Video-Swin-T~\cite{videoswint} as our spatial-temporal visual encoder. We discard the output of its fourth block because of its low spatial resolution, so the output of the third block is regarded as the input visual features of the transformer encoder while the outputs of the earlier blocks are fed into an FPN-like spatial decoder to restore high-resolution visual features. For the linguistic encoder, we utilize the RoBERTa-base model~\cite{roberta} implemented in Hugging Face~\cite{HuggingFace}. 
During training, we follow ~\cite{MTTR} to feed $w=8$ frames around the annotated frame into the model. The input frames are resized such that the height is at least 360 and the width is at most 576. 
They are horizontally flipped with a probability of 0.5. In the transformer part, we utilize 4 encoder blocks and 3 decoder blocks with hidden dimension $D = 256$.
The transformer encoder is modified according to Sec.~\ref{sec:long-short encoder blocks}. 
% The self-attention modules in two of the encoder blocks are replaced by long-short cross-attention modules. 
% The number of attention heads in the transformer encoder and decoder layer is 8. 
The number of object queries is set to 50. The losses weights in Eqn.~\ref{lossfunction} are set to 2, 5, 5, 0.1, for $\lambda_{cls}$, $\lambda_{seg}$,  $\lambda_{lsi}$, and $\lambda_{fbc}$, respectively.
Note that the focal loss in the segmentation loss term $\mathcal{L}_{seg}$ is with a factor 0.4. We set the learning rate as $10^{-4}$ and a batch size of 3. Except for $\lambda_{lsi}$ and $\lambda_{fbc}$, other parameters are set the same as in~\cite{MTTR}.

\noindent
\textbf{LoSh-S}. 
%We also introduce the LoSh built on ReferFormer\cite{ReferFormer} named LoSh-R. 
For LoSh-S, we only emphasize its main differences from LoSh-M. The number of input frames is 5 and the width of each frame is at most 640.  We feed the fused multi-modal features into the deformable transformer, {which has 4 deformable encoder and decoder blocks. 
%The self-attention 
%\miaojing{if you use -, then cross-attention should be using as well} \linfeng{Done. Add '-' for all 'self attention' and  'cross attention' in the paper.} 
%modules in two of the encoder blocks are replaced by the proposed long-short cross-attention modules. 
The number of object queries is set to 5 and the linguistic features of the input text expression are added to each object query as in~\cite{SgMg}. Following~\cite{SgMg}, we pre-train the model on RIS datasets~\cite{refcoco,cocog} and fine-tune it on RVOS datasets. To be consistent with~\cite{SgMg}, we add the same detection loss functions and weighting factors in LoSh-S.
We refer the readers to \cite{SgMg} for more details.

\noindent
\textbf{Ground truth masks for short text expressions}.} 
%\miaojing{REWRITE! CHECK THE REBUTTAL AGAIN!} \linfeng{Done. Remove the 'manually' and emphasize that this process is easy to perform. Put the description of rebuttal in the supplementary.} 
In RVOS datasets, each video clip contains one or multiple target instances, each corresponding with a (long) text expression and mask annotation. For our extracted short text expressions, there exist a few cases (around 10\% in RVOS datasets) when they correspond to multiple instances. 
%Based on current annotations, we can easily collect mask annotations for short text expressions. Specifically, a short text expression of a target instance seldom (approximately 10\%) refer to multiple instances in a video clip. 
In these cases, we simply merge the mask annotations of all the referred instances as the ground truth mask for certain short text expression. 

\subsection{{Comparison with state of the art}}
\label{subsec:compare}
\noindent \textbf{A2D-Sentences.} We first compare our model with state of the art methods on A2D-Sentences. As shown in Tab.~\ref{tab:SOTAcomparison_combined}, our LoSh-M built upon MTTR outperforms the baseline model by a large margin across all metrics. Specifically, LoSh-M shows  +3.1 mAP, +2.7\% Overall IoU and +3.1\% Mean IoU gains compared with MTTR. We also build our LoSh on the previous best-performing method, SgMg. Following~\cite{SgMg}, we first pre-train our LoSh-S on RIS datasets~\cite{refcoco, cocog} and then fine-tune on A2D-Sentences. Our LoSh-S with Video-Swin-T shows clear gains, \eg, +1.5 on mAP, +1.3\% on Overall IoU and +1.2\% on Mean IoU, over SgMg with the same backbone. Given a more powerful backbone (\ie, Video-Swin-B), our proposed LoSh-S consistently outperforms SgMg and other state of the art on all metrics. We provide the results of these models evaluated on JHMDB-Sentences in the supplementary material.

% \noindent \textbf{JHMDB-Sentences.} To evaluate the generalizability of LoSh, we also follow~\cite{MTTR, ReferFormer, SgMg} to evaluate the trained LoSh-M and LoSh-S from A2D-Sentences on JHMDB-Sentences without fine-tuning. As shown in Tab.~\ref{tab:SOTAcomparison_combined}, LoSh-M and LoSh-S also gain massive improvements on all metrics compared to their counterpart baselines. Furthermore, LoSh-S with Video-Swin-B yields even higher results amongst all.

\noindent \textbf{Refer-YouTube-VOS and Refer-DAVIS17.} %We pre-train our LoSh-S$^*$ on RIS datasets and then fine-tune on Refer-YouTube-VOS. 
As shown in Tab.~\ref{tab:SOTAcomparison_combined}, LoSh-S outperforms the baseline SgMg with Video-Swin-T: +1.7 on $\mathcal{J} \& \mathcal{F}$, +1.6 on $\mathcal{J}$ and +1.9 on $\mathcal{F}$. With a stronger backbone (\ie, Video-Swin-B), LoSh-S also exceeds all the methods with a large margin. 

To demonstrate the generalizability of LoSh-S, we follow~\cite{ReferFormer, SgMg} to evaluate the trained LoSh-S from Refer-YouTube-VOS on Refer-DAVIS17 without fine-tuning. As shown in Tab.~\ref{tab:SOTAcomparison_combined}, LoSh-S also gains massive improvements on all metrics compared to its counterpart baseline. 

%To further demonstrate the effectiveness of LoSh, we also provide the results of another LoSh variant on these two datasets in the supplementary material.

%To demonstrate LoSh's compatibility with other RVOS frameworks, we also adapt it to ReferFormer~\cite{ReferFormer}, denoted as LoSh-R. We provide quantitative comparison of LoSh-R$^*$ on Refer-YouTube-VOS and evaluation results of trained LoSh-R$^*$ and LoSh-S$^*$ on Refer-DAVIS17 in the supplementary material. \linfeng{Add analysis and mention that more results about LoSh-R and another RVOS datasets are in supp.}

\subsection{Ablation studies}
\label{ablation study}
In this section, we conduct  ablation studies on A2D-Sentences dataset using our model, LoSh-M.
%to evaluate the effect of the components in our LoSh. The model we select is LoSh-M with details introduced in \ref{implementation} otherwise specify. \linfeng{More ablation studies on xxx can be found in Supplementary.}

\begin{table}
\begin{center}
\begin{tabular}{c|cc|c}
\hline
\multirow{2}{*}{Method} & \multicolumn{2}{c|}{IoU} & \multirow{2}{*}{mAP}  \\ \cline{2-3}
 & Overall& Mean& \\
\hline\hline
Baseline & 70.2 & 61.8 & 44.7\\
LoSh-M w/o Sh & 71.1 & 62.6 & 45.4\\
LoSh-M w/o $\mathcal L_{lsi}$ & 72.3 & 63.8 & 46.1\\
\hline 
LoSh-M w/o CA & 72.5& 64.5& 47.3\\
LoSh-M w/ Inv-CA  &72.5 &64.3 &47.2          \\
LoSh-M w/ Bi-CA &72.8 &64.7 &47.6\\
\hline 
LoSh-M w/ MSh &\textbf{72.9} &64.8 &\textbf{47.8}\\
\textbf{Losh-M (Ours)} & \textbf{72.9} &\textbf{64.9} & \textbf{47.8}\\
% w/o $\mathcal{L}_{lsi}$ & 72.3 \color{red}{(-0.2)} & 63.5 \color{red}{(-1.0)} & 44.5 \color{red}{(-2.7)}\\
% w/o $\mathcal{L}_{fbc}$ & 72.2 \color{red}{(-0.3)} & 64.0 \color{red}{(-0.5)} & 46.8 \color{red}{(-0.4)}\\
% \hline
% Full LoSh-M & 72.5 &64.5 &47.2\\

\hline
\end{tabular}
\end{center}
\setlength{\abovecaptionskip}{-0.3cm}
\setlength{\belowcaptionskip}{-0.3cm}
\caption{Ablation study for long-short text joint prediction.}
\vspace{-6pt}
\label{tab:Ablation for long short}
\end{table}

\setlength{\parskip}{-4pt}
\subsubsection{Long-short text joint prediction} 

We first study the performance of LoSh-M using only the long text expression without the short text expression (LoSh-M w/o Sh), which is equivalent to our baseline MTTR plus the forward-backward visual consistency. 
%\textbf{Baseline + FBC.} 
%Our baseline model is query-based transformer MTTR~\cite{MTTR}. We perform the forward-backward visual consistency loss (FBC) introduced in Sec.~\ref{sec:forward-backward consistency}. 
The result is reported in Tab.~\ref{tab:Ablation for long short}: compared to LoSh-M, LoSh-M w/o Sh has a clear drop on the mAP and IoU, \eg, 2.4 on mAP, 1.8\% on Overall IoU and 2.3\% on Mean IoU.  This indicates the significant benefit of adding short text expression into RVOS.  
\setlength{\parskip}{0pt}

\noindent
\textbf{{Long-short cross-attention.}}
Next, we present the result of LoSh without long-short cross-attention modules, \ie  LoSh-M w/o CA in Tab.~\ref{tab:Ablation for long short}. We observe a 0.5 decrease on mAP and 0.4\% decreases on both Overall IoU and Mean IoU. Recalling that the proposed long-short cross-attention is a unidirectional attention mechanism from $\mathcal F^\text{s}$ to $\mathcal F^\text{l}$, we also offer variants of inverse and bidirectional cross-attentions, denoted by LoSh-M w/ Inv-CA and LoSh-M w/ Bi-CA. The former uses $\mathcal F^\text{l}$ to strengthen $\mathcal F^\text{s}$ while the latter is a combination of both the proposed CA and its inverse version. According to Tab.~\ref{tab:Ablation for long short}, both variants obtain inferior results than LoSh-M with the proposed CA (Ours). 
The result of LoSh-M w/ Inv-CA is even worse than that of LoSh-M without any CA (\ie, LoSh-M w/o CA). 
% Rewritted sentences:
%We suggest that using $\mathcal F^\text{l}$ to reinforce $\mathcal F^\text{s}$ results in a loss of appearance-related information in $\mathcal F^\text{s}$. %The segmentation performance of the model degrades due to the loss of this important auxiliary information. \linfeng{Rewritten here.}
%We suggest that using $\mathcal F^\text{l}$ to reinforce $\mathcal F^\text{s}$ can dilute the appearance-related information in $\mathcal F^\text{s}$. $\mathcal F^\text{s}$ is interfered by action-related information and therefore cannot provide our model with accurate appearance-related information of the target instance which is important for the framework.
These results prove the effectiveness of our proposed unidirectional cross-attention module (see discussion in Sec.~\ref{sec:long-short encoder blocks}).

\noindent
\textbf{Long-short predictions intersection loss.} Next, we study the proposed long-short predictions intersection loss $\mathcal L_{lsi}$ by presenting a variant of LoSh-M without  using $\mathcal L_{lsi}$, \ie, LoSh-M w/o $\mathcal L_{lsi}$. The long and short text expressions are still used and long-short cross-attention still aggregates useful information from $ \mathcal F^\text{s}$ to $\mathcal F^\text{l}$. Tab.~\ref{tab:Ablation for long short} shows the result. We observe a 1.7 decrease on mAP from LoSh-M to LoSh-M w/o $\mathcal L_{lsi}$. Without using $\mathcal L_{lsi}$, the model can not regulate the predicted masks from the long and short text expressions.

%prove the importance of utilizing short text expressions. 

%Despite there is no explicit interaction between the two predictions from long and short text expressions, $\mathcal L_{lsi}$  
%The model is simply trained with the loss of predictions for long-short text expressions as well as the FBC. The model shows inferior performance on the important metric, mAP. I
%t results from the reason that the LoSh-M without Long-short prediction intersection loss cannot associate the predicted pixels for the long and short text expressions.
% Although the segmentation loss is backpropagated through the predicted pixels, the model cannot discover useful information from the input long-short text expressions, which leads to failure in the alignment of visual and textual features.

%\noindent
%\textbf{Long-short prediction intersection Loss(LSI).}  We add LSI to the model and gain significant improvements in mAP (+2.7). It means that our proposed LSI can discover and regulate the inclusion relationship between the long and short text expressions and that between their corresponding predicted pixels. LSI can force the predictions for the long text expressions  to pay more attention to the predicted pixels in the short ones.

\noindent
\textbf{Short text expressions generation.} As mentioned in Sec.~\ref{sec:long-short text expression}, the short text expressions can be manually or automatically extracted from long text expressions.
% or automatically generated by a machine.  
For the latter, we feed the long text expressions into a part-of-speech tagger provided by the NLTK library~\cite{NLTK}: we can extract the words before the first verb (including `is') predicted by the tagger as short text expressions. We denote this variant of using machine-generated short text expression as LoSh-M w/ MSh and report its result in Tab.~\ref{tab:Ablation for long short}. It produces almost the same mAP and relatively closed IoU as the original LoSh-M. 
{Statistically, we verify that 98.7\% of machine-generated short expressions are identical to our manually extracted ones.}
It demonstrates the flexibility of our LoSh and its potential to transfer to larger datasets.

\setlength{\parskip}{-4pt}
\subsubsection{Forward-backward visual consistency}
We first present the result of LoSh without using the forward-backward visual consistency loss, \ie, LoSh-M w/o $\mathcal L_{fbc}$ in Tab.~\ref{tab:Ablation_optical-flow}. We observe a 0.5 decrease on mAP and 0.4\% as well as 0.5\% decreases on Overall IoU and Mean IoU, compared to LoSh-M. 
%LoSh-M without FBC shows a -0.4 mAP loss. The IOU metrics are also degraded. 
{Merely integrating $\mathcal L_{fbc}$ to  baseline MTTR (LoSh-M w/o Sh in Tab.~\ref{tab:Ablation for long short}) also leads to clear improvement, illustrating the effectiveness of $\mathcal L_{fbc}$.} 

\setlength{\parskip}{0pt}

%fbc
% In terms of our visual innovation  $\mathcal{L}_{fbc}$, the third row in Tab.~\ref{tab:Ablation} indicates that the forward-backward visual consistency loss can enhance the temporal consistency of features across multiple frames and bring a +0.4 mAP gain.

\begin{table}
\begin{center}
\begin{tabular}{c|cc|c}
\hline
\multirow{2}{*}{Method} & \multicolumn{2}{c|}{IoU} & \multirow{2}{*}{mAP}  \\
\cline{2-3}
 & Overall& Mean& \\

\hline\hline
LoSh-M w/o $\mathcal{L}_{fbc}$ & 72.5 & 64.4  & 47.3 \\

LoSh-M w/ $\mathcal{L}_{ofbc}$ & 72.6 &64.7 &47.5\\
LoSh-M w/ $\mathcal{L}_{mfbc}$ & {72.8}&\textbf{64.9} &47.7\\
\textbf{LoSh-M (Ours)} & \textbf{72.9} &\textbf{64.9} &\textbf{47.8}\\

\hline
\end{tabular}
\end{center}
\setlength{\abovecaptionskip}{-0.3cm}
\setlength{\belowcaptionskip}{-0.2cm}
\caption{Ablation study for forward-backward visual consistency.}
\vspace{-6pt}
\label{tab:Ablation_optical-flow}
\end{table}

\noindent
\textbf{Directed consistency.} In Sec.~\ref{sec:forward-backward consistency}, we warp the features from multiple adjacent frames to the annotated frame. On the contrary, we can also warp the feature of the annotated frame to its temporal neighbors and compute the consistency loss. We denote this variant of using opposite directions as  
$\mathcal L_{ofbc}$ and show the result in Tab.~\ref{tab:Ablation_optical-flow}. The result can also be improved this way yet it is slightly inferior to $\mathcal L_{fbc}$, as we care more about the annotated frame. Besides, we also try to combine $ \mathcal L_{ofbc}$ and $\mathcal L_{fbc}$ in the consistency, and denote this result of using mutual warping as $\mathcal L_{mfbc}$, the performance is basically the same to using the original $\mathcal L_{fbc}$, suggesting the mutual warping is unnecessary.

\begin{table}[t]
%\scriptsize
\begin{center}
\begin{tabular}{c|cccc}
\hline
% {Method} & GFLOPs  & Train Time (ms) &  Inf Time (ms) & Flow Time (ms) \\
{Method} & FLOPs  & $\mathcal{T_\text{train}}$ & 
$\mathcal{T_\text{infer}}$ \\

\hline\hline

MTTR  & 238.9G & 141ms & 79ms \\

LoSh-M & 253.2G & 153ms & 80ms \\

\hline
\end{tabular}
\end{center}
\setlength{\abovecaptionskip}{-0.3cm}
\setlength{\belowcaptionskip}{-0.2cm}
\caption{Computation cost, training and inference time for a sample. The resolution of the input frames is $320\times576$. %The length of the long text expression is 7 while that of the short one is 3. 
%The  $\mathcal{T_\text{of}}$ represents the time cost of calculation of optical flow during training.
}
\vspace{-6pt}
\label{tab:computation cost}
\end{table}

\subsection{Analysis of computation cost}
\label{exp:cost}
In Tab.~\ref{tab:computation cost}, we provide the  computation cost (\ie, FLOPs) and training time of LoSh. %We only present the comparison between LoSh-M and its baseline MTTR while a similar observation goes between LoSh-S and SgMg.
Although LoSh introduces an additional branch to generate a mask for the short text expression, it actually incurs only a little additional computation cost (5\% of that of the baseline) and training time (8\% of that of the baseline). 
The reasons are that 1) short text expressions are on average one-third of the length of long ones; 2) 
around 90\% of computation budgets are consumed by the visual encoder which is also only performed once in LoSh.
%\miaojing{why?} \linfeng{Done.}
Last, we also measure the time cost of optical flow calculation which is only performed during training.
The calculation of optical flow consumes CPU time (74ms). It can happen in parallel when the GPU is processing the previous sample (153ms) or even offline, hence not reducing the overall training efficiency. {Finally, the inference time is also provided, our LoSh-M basically consumes the same time to its baseline.}

%and training time of LoSh given one input sample. The time to calculate optical flows for an input video clip is also illustrated. The experimental setup is the same as that mentioned in Sec.~\ref{implementation}. 

%\noindent \textbf{Computation cost of LoSh.} 

\begin{figure}
\begin{center}
\includegraphics[width=0.9\linewidth]{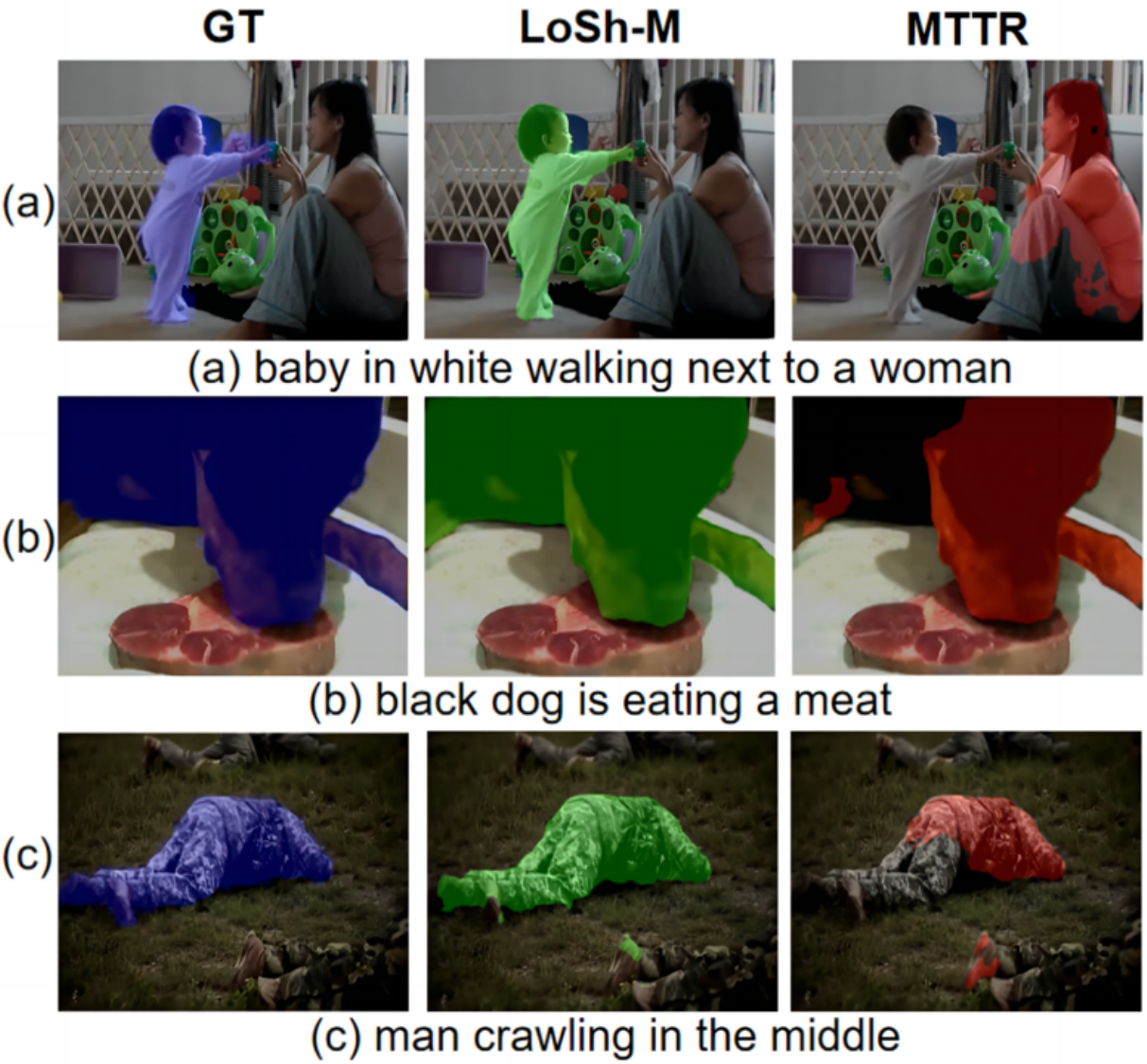}
\end{center}
\setlength{\abovecaptionskip}{-0.3cm}
\setlength{\belowcaptionskip}{-0.2cm}
   \caption{Qualitative comparison between LoSh-M and MTTR {on A2D-Sentences}. LoSh-M generates reasonable  predictions while MTTR predicts incorrect or partial ones compared to ground truth.}
\vspace{-6pt}
\label{fig:qualitative_comparison}
\end{figure}

\subsection{Qualitative results}
%\linfeng{Here I move the qualitative comparison in the main body and put the visualization in the supplementary.}
We also provide qualitative comparison between LoSh-M and baseline MTTR.~Fig.~\ref{fig:qualitative_comparison} displays the cases where LoSh-M generates reasonable predictions while MTTR predicts incorrect or partial ones.
{This improvement is attributed to our model's enhanced attention on the appearance information. It can mitigate excessive impact of actions (\eg, `eating') or relations (\eg, `in the middle').}
%In each case, MTTR produces unreasonable prediction due to the excessive attention to the action- or relation-related information \textcolor{red}{(\eg, `next to a woman', `eating')} and the neglect of the appearance-related information \textcolor{red}{(\eg,`baby in white', `black dog')} of the target instance in the input text expression. Under the proposed long-short text joint prediction framework, LoSh can produce more reasonable predictions compared to its counterpart baseline. 
We provide more qualitative results in the supplementary material. 

\section{Conclusion}
In this work, we propose LoSh, the long-short text joint prediction network for referring video object segmentation. 
We generate short text expressions from original long text expressions in RVOS and propose a long-short cross-attention module and a long-short predictions intersection loss to regulate the segmentation on the referred instance.
Besides, a forward-backward visual consistency loss is also injected into LoSh to warp between the features of adjacent frames for visual consistency.  
%into state of the art single-stage RVOS frameworks. 
Our proposed method can be easily plugged into many RVOS frameworks and brings no significant extra time during inference. Specifically, we build our method on top of two state of the art RVOS pipelines~\cite{MTTR, SgMg}, and achieve significant improvements over the previous best-performing methods. 

{
    \small
    \bibliographystyle{ieeenat_fullname}
    \bibliography{RefList}
}


\end{document}